\documentclass[a4paper]{article}

\usepackage{authblk}
\usepackage[utf8]{inputenc}
\usepackage[T1]{fontenc}
\usepackage{pslatex}
\usepackage[english]{babel}
\usepackage{amsmath,amssymb,amsfonts}
\usepackage{amsthm}
\usepackage{graphicx}
\usepackage{booktabs}
\usepackage{paralist}
\usepackage{url}
\usepackage{microtype}
\usepackage{cite}
\usepackage{hyperref}

\newcommand{\N}{\mathbb{N}}

\newcommand{\F}{\mathbb{F}}

\theoremstyle{plain}
\newtheorem{lemma}{Lemma}
\theoremstyle{definition}
\newtheorem{definition}{Definition}
\newtheorem{problem}{Problem}

\providecommand{\keywords}[1]{\noindent\textbf{\textit{Keywords }} #1}

\begin{document}

\title{A Matter of Representation: Evolving Partial Spreads for Bent Functions}

\author[1,2]{Claude Carlet}
\author[3]{Marko \DH urasević}
\author[3]{Domagoj Jakobovic}
\author[4]{Luca Manzoni}
\author[5]{Luca Mariot}
\author[3,6]{Stjepan Picek}

\affil[1]{{\normalsize Department of Mathematics, Universit\'{e} Paris 8, 2 rue de la libert\'{e}, 93526 Saint-Denis Cedex, France}}

\affil[2]{{\normalsize University of Bergen, Bergen, Norway}\protect\\
	{\small \texttt{claude.carlet@gmail.com}}}

\affil[3]{{\normalsize Faculty of Electrical Engineering and Computing, University of Zagreb, Unska 3, Zagreb, Croatia}\protect\\
		{\small \texttt{\{marko.durasevic, domagoj.jakobovic\}@fer.hr}}}

\affil[4]{{\normalsize Department of Mathematics, Informatics and Geosciences, University of Trieste, Via Alfonso Valerio 12/1, 34127 Trieste, Italy}\protect\\
	{\small \texttt{lmanzoni@units.it}}}

\affil[5]{{\normalsize Semantics, Cybersecurity and Services Group, University of Twente, Drienerlolaan 5, 7522 NB Enschede, The Netherlands}\protect\\
	{\small \texttt{l.mariot@utwente.nl}}}

\affil[6]{{\normalsize Digital Security Group, Radboud University, Postbus 9010, 6500 GL Nijmegen, The Netherlands}\protect\\
	{\small \texttt{stjepan.picek@ru.nl}}}

\maketitle

\begin{abstract}
\noindent
Partial spread ($\mathcal{PS}$) bent functions are Boolean functions of maximum nonlinearity built from families of subspaces that pairwise intersect only in the null vector. We design an Evolution Strategy (ES) to construct such partial spread families, comparing two encodings: one based on generator matrices of the subspaces, and one based on spread sets. With the latter, combined with a ruin-and-rebuild operator, the ES finds $\mathcal{PS}$ functions up to $n=10$ variables. An ablation study shows that mutation is necessary, while the population size is not, and that a restarted look-ahead greedy heuristic outperforms the ES on the hardest instance. We investigate this finding with a landscape analysis of the stagnation points, and we also introduce a quality-diversity variant of the ES which finds $\mathcal{PS}$ functions of low algebraic degree for 8 variables. Interestingly, the bent functions evolved by the ES have higher 2-ranks and mostly higher degrees than those obtained by a baseline Genetic Programming optimizer.

\end{abstract}

\keywords{Bent functions, Partial spreads, Evolution strategies, Representation, Fitness landscape analysis}

\section{Introduction}\label{sec:intro}
In the mid-19th century, Reverend Thomas Kirkman posed the following combinatorial puzzle in \emph{The Lady's and Gentleman's Diary}~\cite{kirkman1850}: given fifteen young ladies in a school, how can they walk out three abreast for seven days in such a way that no two of them walk twice in the same row? This puzzle later became known as \emph{Kirkman's Schoolgirl Problem}~\cite{colbourn2019}. An elegant solution, which goes back at least to Rao~\cite{rao46}, identifies the fifteen schoolgirls with the nonzero strings of four bits, and lets any three girls walk in a row only if the XOR of their strings is the all-zero string. Each day then partitions the fifteen girls into five such rows, and seven days with no rows in common provide a solution to the riddle~\cite{brown2009}.

From a geometric standpoint, each row (augmented with the all-zero string) is closed under XOR, and it is thus a 2-dimensional subspace of $\F_2^4$. Any subset of rows on a given day is a family of subspaces that intersect only in the null vector, which is called a \emph{partial spread}. This construction yields another interesting combinatorial object: the six girls (or more precisely, their 4-bit strings) in any two rows on a single day define the support of a \emph{bent function}.

Bent functions are those Boolean functions that are as far as possible from linear ones, and they have several applications in cryptography, coding theory, and sequence design~\cite{carlet21}. The example above is the smallest instance of a general construction introduced by Dillon~\cite{dillon74}, namely the \emph{partial spread class} $\mathcal{PS}^-$: for $n=2m$ variables, the union of $2^{m-1}$ $m$-dimensional subspaces of $\F_2^n$ with trivial pairwise intersection gives the support of a bent function, once the null vector is removed. The class $\mathcal{PS}^+$ instead takes $2^{m-1}+1$ subspaces and keeps the null vector in the support. The union of $\mathcal{PS}^-$ and $\mathcal{PS}^+$ gives the partial spread bent functions class $\mathcal{PS}$. To obtain a function in this class, one must first find an appropriate partial spread, and exhaustive search quickly becomes infeasible even for small numbers of variables. For instance, with $n=10$ one has to choose 16 subspaces among more than $10^8$ candidates for a $\mathcal{PS}^-$ function, turning what is a brain teaser for fifteen schoolgirls into a hard combinatorial search problem.

So far, the main approaches in the literature to design bent functions---whether in the partial spread class or not---rely on \emph{algebraic constructions}~\cite{carlet21} or metaheuristic optimization algorithms~\cite{djurasevic23}. The former give complete control over the class of bent functions that can be obtained, but less variability. The latter provide, in principle, more diversity, since they blindly search the space of truth tables (or equivalent representations of Boolean functions, like syntactic trees), guided only by a fitness function that usually measures the nonlinearity. On the other hand, with metaheuristics it is more difficult to enforce certain properties on the resulting bent functions, such as being defined by a partial spread. A third option is to constrain the metaheuristic to search for the combinatorial object underlying a construction, rather than for the function itself.

The aim of this paper is to investigate to what extent Evolutionary Algorithms (EAs) can be used to evolve bent functions in the partial spread class. Since a subspace can be encoded by a generator matrix---like a linear code---our idea starts from the Evolution Strategy (ES) of~\cite{carlet23} to evolve binary linear codes. However, a direct adaptation of this approach to optimize families of $m\times 2m$ matrices with trivial pairwise intersection already fails at $n=10$ variables. We therefore introduce a \emph{spread-set encoding}, where two fixed subspaces (denoted as $E_0$ and $E_\infty$) always belong to the family. Every remaining member of the family is then the graph of a linear map described by an $m\times m$ matrix, and two such subspaces have trivial intersection if and only if the difference of their matrices is invertible. Since each column of the difference depends only on the same column of the two matrices, a conflicting subspace can be removed and rebuilt in an exact way, instead of being modified one vector at a time as in~\cite{carlet23}.

The contributions of this paper are as follows:
\begin{compactitem}
	\item \emph{Representation.} We compare the performance of the ES with the subspace and spread-set encodings. With $n=8$ variables, the ES always converges under both encodings. For $n=10$, the subspace encoding always fails, while the spread-set encoding combined with the exact ruin-and-rebuild operator achieves a full success rate for the $\mathcal{PS}^-$ target and a 74\% success rate for $\mathcal{PS}^+$. Neither of the two succeeds alone.
	
	\item \emph{Components and Baselines.} We perform an ablation study on the components of the ES. Mutation turns out to be crucial, while the population size has no significant effect. Moreover, a restarted greedy heuristic with look-ahead on the same encoding has a higher success rate than any ES variant on the hardest instance ($n=10$ with the $\mathcal{PS}^+$ target).
	
	\item \emph{Landscape Analysis.} We measure the number of matrices that need to be replaced to reach a partial spread from the points where the ES stagnates. For $n=10$, this number is mostly 4 for the $\mathcal{PS}^-$ target and mostly 7 for $\mathcal{PS}^+$, consistent with the performance of the ES. For $n=12$, neither the ES nor the greedy heuristic finds a solution within 24 hours.
	
	\item \emph{Cryptographic Properties.} We add the algebraic degree as a second optimization objective, to search for $\mathcal{PS}^+$ functions with degree smaller than $n/2$. For $n=8$, a quality-diversity variant of the ES finds such functions more reliably than the greedy heuristic, while for $n=10$ no such function is found. Further, we analyze the functions obtained for $n=10$ in terms of \emph{Extended Affine (EA-) equivalence}: all distinct solutions are pairwise inequivalent, and their 2-ranks show that they are inequivalent to bent functions in the Maiorana--McFarland class and to partial spread functions from the Desarguesian spread. Conversely, the 2-ranks of bent functions evolved by Genetic Programming~\cite{picek16} all lie within the Maiorana--McFarland bound.
\end{compactitem}

\section{Background}
\label{sec:background}
This section recalls the definitions of Boolean functions used in the paper. The reader is referred to~\cite{carlet21} for a more complete treatment of the subject.

\subsection{Boolean Functions and Their Representations}
\label{subsec:boolean}
In what follows, we denote by $\F_2 = \{0,1\}$ the finite field with two elements, with XOR and AND of two bits as the field operations, while $\F_2^n$ denotes the $n$-dimensional vector space over $\F_2$, equipped with the \emph{dot product} $x \cdot y = \bigoplus_{i=1}^n x_iy_i$ and the Hamming distance $d_H(x,y) = |\{i: x_i \neq y_i\}|$ for all $x, y \in \F_2^n$. A Boolean function is a mapping of the form $f: \F_2^n \to \F_2$. The \emph{truth table} of $f$ is the unique $2^n$-bit vector $\Omega_f = (f(0,\ldots,0), f(0,\ldots, 1), \ldots, f(1,\ldots, 1))$ that lists all outputs of $f$ in lexicographic order of the inputs. The \emph{support} of $f$ is defined as $\mathrm{supp}(f) = \{x \in \F_2^n: f(x) \neq 0\}$. The \emph{Walsh--Hadamard transform} of $f: \F_2^n \to \F_2$ is defined for all $a \in \F_2^n$ as $W_f(a) = \sum_{x \in \F_2^n} (-1)^{f(x) \oplus a\cdot x}$, and the coefficient $W_f(a)$ quantifies the correlation between $f$ and the linear function $a\cdot x$. The \emph{nonlinearity} of $f$ is the minimum Hamming distance between the truth table of $f$ and those of the affine functions $\mathcal{A}_n = \{a\cdot x \oplus b: a \in \F_2^n, b \in \F_2\}$, and it equals $\mathrm{nl}_f = 2^{n-1} - \frac{1}{2}\max_{a \in \F_2^n}\{|W_f(a)|\}$. \emph{Bent functions} are defined as those functions $f:\F_2^n \to \F_2$ such that $W_f(a) = \pm 2^{\frac{n}{2}}$ for all $a \in \F_2^n$, and they reach the highest possible nonlinearity of $2^{n-1} - 2^{\frac{n}{2}-1}$ (also called the \emph{covering radius bound}). Bent functions exist only when $n=2m$ for $m \in \N$, since the Walsh--Hadamard coefficients must be integers. The \emph{Algebraic Normal Form} (ANF) of $f: \F_2^n \to \F_2$ is the multivariate polynomial in the quotient ring $\F_2[x_1,\ldots, x_n]/(x_1^2 \oplus x_1, \ldots, x_n^2 \oplus x_n)$ defined as $P_f = \bigoplus_{I \in \mathcal{P}([n])} a_I \prod_{i \in I} x_i$, where $\mathcal{P}([n]) = 2^{[n]}$ is the power set of $[n] = \{1,\ldots, n\}$, and $a_I \in \F_2$ is the coefficient of the monomial defined by the subset $I \in \mathcal{P}([n])$. The \emph{algebraic degree} of $f$ is the size of the largest subset $I$ for which $a_I \neq 0$. Rothaus~\cite{rothaus76} proved that the algebraic degree of any bent function of $n > 2$ variables is at most $n/2$. Two functions $f, g$ are \emph{Extended-Affine (EA-)equivalent} if there exists a linear invertible function $A: \F_2^n \to \F_2^n$, two vectors $u, v \in \F_2^n$ and a constant $c \in \F_2$ such that $g(x) = f(A(x) \oplus u) \oplus (v \cdot x) \oplus c$ for all $x \in \F_2^n$. EA-equivalence preserves bentness, and it also preserves the algebraic degree when the latter is at least 2.

\subsection{Partial Spread Bent Functions}
\label{subsec:psbent}
Among all constructions of bent functions described in the literature, the one based on partial spreads, originally introduced by Dillon~\cite{dillon74}, is especially interesting, since it is defined by a combinatorial object. Given $n=2m$, a \emph{partial spread} is a family $P = \{E_1, E_2, \ldots, E_t\}$ of $m$-dimensional subspaces of $\F_2^n$, such that $E_i \cap E_j = \{\underline{0}\}$ for each $i \neq j$, that is, each pair of subspaces in $P$ has trivial intersection. One has a \emph{full spread} when $t = 2^m+1$ (i.e., the family covers the entire space $\F_2^{n}$). In the solution of Kirkman's schoolgirl problem described in Section~\ref{sec:intro}, each day is a full spread of $\F_2^4$. For bent functions, we are interested in partial spreads where $t=2^{m-1}$ or $t=2^{m-1}+1$, as per the following definition:
\begin{definition}
Let $n=2m$ and $P = \{E_1, E_2, \ldots, E_t\}$ be a partial spread of $m$-dimensional subspaces, with $t = 2^{m-1}$ (respectively, $t = 2^{m-1}+1$). A function $f: \F_2^n \to \F_2$ is of type $\mathcal{PS}^-$ (respectively, $\mathcal{PS}^+$) if its support is defined as:
\begin{equation}
\label{eq:supp}
\mathrm{supp}(f) = \bigcup_{E_i \in P} (E_i \setminus \{\underline{0}\}) \enspace \left( \textrm{respectively, } \mathrm{supp}(f) = \bigcup_{E_i \in P} E_i \right) \enspace .
\end{equation}
\end{definition}
Dillon~\cite{dillon74} proved that every function of type $\mathcal{PS}^-$ or $\mathcal{PS}^+$ is bent. A partial spread $P$ is \emph{extendable} if there exists an additional $m$-dimensional subspace that trivially intersects all members of $P$. The degree of any $\mathcal{PS}^-$ function is exactly $m$, and lower degrees are possible only for $\mathcal{PS}^+$ functions whose partial spreads are \emph{non-extendable}~\cite{weng07}.

The \emph{Desarguesian (full) spread} is defined by identifying $\F_2^n$ with $\F_{2^m}^2$ as follows: $DS = \{E_a \subseteq \F_{2^m}^2: a \in \F_{2^m}\} \cup \{E_{\infty}\}$, where $E_a = \{(x, ax) \in \F_{2^m}^2: x \in \F_{2^m}\}$ and $E_\infty = \{(0, y) \in \F_{2^m}^2 : y \in \F_{2^m}\}$. Taking $2^{m-1}$ subspaces from $DS$ gives the partial spread bent functions in the class $\mathcal{PS}_{ap}$ (where $ap$ stands for \emph{affine plane}), and taking $2^{m-1}+1$ subspaces gives their complements.

The 2-rank of a function $f: \F_2^n \to \F_2$ is the rank over $\F_2$ of the $2^n\times 2^n$ matrix $A_f(x,y) = f(x \oplus y)$ for all $x, y \in \F_2^n$, and if two bent functions $f,g$ have different 2-ranks, then they are EA-inequivalent. Weng et al.~\cite{weng07} computed lower and upper bounds on the 2-ranks of $\mathcal{PS}_{ap}$ functions and of functions in the \emph{Maiorana--McFarland} (MM) class~\cite{mcfarland73}. In particular, the 2-rank of an MM function is at most $2^{m+1}-2$, while the 2-rank of a $\mathcal{PS}_{ap}$ function lies between $2^{m+1}-2$ and $\sum_{r=0}^{m} \binom{m}{r} 2^{\min\{r, m-r\}}$. Thus, if the 2-rank of a bent function lies outside these bounds, it is not EA-equivalent to any function in the respective class. The 2-rank is commonly used to establish the inequivalence to the MM and $\mathcal{PS}_{ap}$ classes of bent functions generated by new constructions (see, e.g.,~\cite{gadouleau23}).

\section{Related Work}
\label{sec:related}
This section surveys the main works concerning metaheuristic algorithms for designing bent functions, referring the reader to~\cite{djurasevic23} for a broader treatment.

To the best of our knowledge, Fuller et al.~\cite{fuller03} were the first to explicitly consider a metaheuristic approach for constructing bent functions. Despite the title mentioning an ``Evolutionary Algorithm'', the paper describes a randomized local search algorithm that explores the space of algebraic normal forms, starting from a known bent function to find new ones with different degrees. Stanica et al.~\cite{StanicaMC04} adapted the simulated annealing approach of~\cite{clark03} to search for bent functions in the rotation-symmetric class, i.e., those functions that are invariant under cyclic shifts of the inputs. Later, Hrbacek and Dvorak~\cite{hrbacek14} applied Cartesian Genetic Programming (CGP) to evolve feed-forward circuits that were used to synthesize the truth table of Boolean functions, with the objective of maximizing their nonlinearity. Their approach showed that they could routinely design bent functions up to $n=16$ variables. Picek et al.~\cite{picek16} performed a comparison of different EAs---namely ES, Genetic Algorithms (GA), Genetic Programming (GP) and CGP---to evolve Boolean functions of $n=8$ variables with different cryptographic profiles. When maximizing the nonlinearity only, the symbolic encodings of GP and CGP found bent functions reliably, while the truth table-based representations of ES and GA never did within the given fitness budget. A similar difficulty was observed by Manzoni et al.~\cite{manzoni20}, who restricted the GA to search only for truth tables with the Hamming weight of bent functions by employing weight-preserving crossover and mutation.

A parallel line of work investigated EAs to evolve variants of bent functions, including bent quaternary functions~\cite{picek18}, hyper-bent functions~\cite{mariot19}, self-dual bent functions~\cite{carlet24a}, homogeneous bent functions~\cite{carlet26a} and negabent functions~\cite{carlet26b}.

The works reviewed so far evolve bent functions with a \emph{direct search}, e.g., directly searching the spaces of truth tables or syntactic trees. A different approach considers the use of EAs to evolve \emph{secondary constructions}, i.e., formulae that combine existing bent functions to obtain new ones on a larger number of variables. In this thread, Picek and Jakobovic~\cite{picekJ16} used GP to evolve secondary constructions for bent functions in $n+2$ variables, showing that it scales to $n=20$ variables, where a direct search GP with the same budget stops at $n=12$. Mariot et al.~\cite{mariot22} developed an ES algorithm to evolve quadratic bent functions via a secondary construction based on Cellular Automata (CA); the resulting functions all belong to the same class under EA-equivalence.

Contrary to the works mentioned above, our paper does not use EAs to find bent functions by direct search or by evolving secondary constructions; instead, we tackle the optimization of a combinatorial object that underlies a primary construction, namely a partial spread, so that any obtained bent function is in the $\mathcal{PS}$ class. This idea was sketched as an open challenge in~\cite{djurasevic23}. Further, excluding~\cite{fuller03}, none of the works reviewed above analyzes the functions up to EA-equivalence, which we address by means of the 2-rank.

\section{Algorithms}
\label{sec:methods}
In this section, we describe the main components of the ES that we use to evolve partial spread bent functions, including the fitness function, the underlying encodings and the variation operators. We also describe the heuristic baselines we compare against our ES.

\subsection{Search Space and Fitness}
\label{sec:fitness}
Since a partial spread is a family $P$ of vector subspaces, the main idea of this paper is to turn the construction of $P$ into a combinatorial optimization problem, which we formally state as follows:
\begin{problem}
\label{prob:stat}
    Let $n=2m$. Find a set $P = \{E_1, E_2, \ldots, E_t\}$ of $m$-dimensional subspaces of $\F_2^{n}$ with pairwise trivial intersection, where $t=2^{m-1}$ for $\mathcal{PS}^-$ functions and $t= 2^{m-1}+1$ for $\mathcal{PS}^+$ functions.
\end{problem}
The number of $m$-dimensional subspaces of $\F_2^{n}$ equals the \emph{Gaussian binomial coefficient} $\binom{n}{m}_2$~\cite{mullen13}. A candidate solution for Problem~\ref{prob:stat} is a set $P$ of $t$ such subspaces that may intersect non-trivially. This means that the size of the search space $\mathcal{S}_m$ for $n=2m$ variables in Problem~\ref{prob:stat} is:
\begin{equation}
    \label{eq:space}
    |\mathcal{S}_m| = \binom{\binom{n}{m}_2}{t} \enspace ,
\end{equation}
which is infeasible to enumerate exhaustively already for $n=8$ variables, since in that case $\mathcal{S}_m$ contains approximately $6.6 \cdot 10^{37}$ candidate families for $t=8$.

Denoting by $d_{ij} = \dim(E_i \cap E_j)$ the dimension of the intersection of two subspaces, the fitness for a candidate $P = \{E_1,\ldots, E_t\}$ is defined as:
\begin{equation}
    \label{eq:fitness}
    \mathrm{\emph{fit}}(P) = \sum_{i<j} \left(2^{d_{ij}} - 1\right) \enspace .
\end{equation}
Thus, $\mathrm{\emph{fit}}$ computes the number of nonzero vectors shared by each pair and sums them. The fitness function must be minimized, and an optimal solution $P$ corresponds to a family with $\mathrm{\emph{fit}}(P)=0$, i.e., all subspaces in it have pairwise trivial intersection. Remark that when only a single subspace is changed in $P$, only $t-1$ intersections need to be recomputed. A \emph{conflicting pair} is a pair $E_i, E_j \in P$ such that $d_{ij}>0$; further, a \emph{conflicting subspace} $E_i \in P$ is a subspace such that there exists $E_j \in P$ with $d_{ij}>0$.

Besides Problem~\ref{prob:stat}, in this work we are also interested in finding $\mathcal{PS}^+$ functions with degree smaller than $m$. To this end, we define an additional objective as follows. Let $T(f)$ denote the number of monomials of degree $m$ in the ANF of a $\mathcal{PS}^+$ function $f$, which is at most $\binom{n}{m}$. Then, $T(f) = 0$ if and only if the degree of $f$ is smaller than $m$. A first way to combine $T(f)$ with $\mathrm{\emph{fit}}$ is lexicographic: first we minimize $\mathrm{\emph{fit}}$ to reach a partial spread of a $\mathcal{PS}^+$ function, and then we minimize $T(f)$ to get a degree below $m$. A second way, based on quality diversity, is described in Section~\ref{sec:es}.

\subsection{Solution Encodings}
\label{subsec:encodings}

\paragraph{Subspace Encoding.} Our first encoding builds directly on the approach of~\cite{carlet23} for binary linear codes. A candidate solution is represented as a family of $t$ binary matrices $P=\{M_1, \ldots, M_t\}$, each of size $m\times 2m$ and rank $m$, where the rows of $M_i$ form a basis of the subspace $E_i$. Given $M_i, M_j \in P$, remark that $d_{ij} = 2m - \mathrm{rank}(M_{ij})$, where $M_{ij}$ represents the square $2m \times 2m$ matrix obtained by stacking $M_i$ on top of $M_j$.

\paragraph{Spread-set Encoding.} In what follows, we write each vector of $\F_2^n$ as a pair $(x,y)$ with $x, y \in \F_2^m$. The second encoding is based on the following fact in finite geometry (proved in~\cite{bruck64} for spreads, with an argument that also applies to partial spreads): any two subspaces of a partial spread can be mapped by an invertible linear map to $E_0 = \{(x, \underline{0})\}$ and $E_\infty = \{(\underline{0}, y)\}$, and this map preserves the nonlinearity, algebraic degree and 2-rank of the corresponding bent function. Thus, $E_0$ and $E_\infty$ can be fixed in the family without loss of generality. Moreover, any $m$-dimensional subspace that intersects $E_{\infty}$ trivially is the graph $E_A = \{(x, Ax): x \in \F_2^m\}$ of a linear function defined by an $m\times m$ matrix $A$. This gives the following result for the dimension of the intersection of two subspaces:
\begin{lemma}
    \label{lm:dim}
    Let $E_A = \{(x, Ax): x \in \F_2^m\}$ and $E_B = \{(x, Bx): x \in \F_2^m\}$ be two subspaces respectively defined by the $m\times m$ matrices $A$ and $B$. Then, $\dim(E_A \cap E_B) = m - \mathrm{rank}(A - B)$.
\end{lemma}
\begin{proof}
    A vector $(x,y) \in \F_2^{2m}$ belongs to $E_A \cap E_B$ if and only if $y = Ax = Bx$, i.e., $(A - B)x = \underline{0}$. Thus, $x \mapsto (x, Ax)$ is a bijection between $\ker(A - B)$ and $E_A \cap E_B$, and the result follows from the rank--nullity theorem.
\end{proof}
The main consequence of Lemma~\ref{lm:dim} is that $E_A$ and $E_B$ have trivial intersection if and only if the difference $A-B$ of their matrices is invertible.\footnote{Notice that, since we are working over $\F_2$, $A - B$ equals $A\oplus B$ (i.e., the element-wise XOR of the matrices).} In particular, if $B$ is the zero matrix, the intersection $E_A \cap E_0$ is trivial if and only if $A$ is invertible, while $E_A \cap E_\infty$ is trivial for any matrix $A$.

Under this encoding, the genotype of a candidate solution thus becomes a family of $t-2$ $m\times m$ binary matrices $\{A_1, \ldots, A_{t-2}\}$, which stands for $P = \{E_\infty, E_0, E_{A_1}, \ldots, E_{A_{t-2}}\}$, and Problem~\ref{prob:stat} can be reformulated as finding $t-2$ invertible matrices whose pairwise differences are all invertible. This encoding imposes the graph form on all candidates, including those that have conflicts, and the fitness function remains the same, since $d_{ij}$ can be obtained by Lemma~\ref{lm:dim}.

The spread-set encoding is basically equivalent to the subspace encoding described above, where the generator matrix is in \emph{systematic form} and two members are fixed. Contrary to the subspace encoding, each subspace $E_A$ is represented by exactly one matrix. Moreover, since our main variation operators keep every matrix invertible (see Section~\ref{sec:operators}), the number of possible candidates for each position in the family drops from $\binom{n}{m}_2$ subspaces to the number of invertible $m\times m$ matrices: for example, for $m=5$, from $109\,221\,651$ to $9\,999\,360$.

\subsection{Variation Operators}
\label{sec:operators}
With the subspace encoding, our ES algorithm uses the following two operators:

\begin{compactitem}
\item \emph{Row replacement:} this operator is taken from~\cite{carlet23}. Given an $m \times 2m$ generator matrix $M_i$ in a candidate family $P$, the operator randomly picks one of its rows and removes it. The reduced matrix now spans a subspace $W_i$ of dimension $m-1$. Then, a random vector $v$ is drawn from $\F_2^n \setminus W_i$, i.e., the complement of $W_i$, so that adjoining it to the reduced matrix gives a new matrix $M'_i$ of rank $m$. Each matrix/row pair is selected with probability $1/(tm)$, and if none is selected, one replacement is made anyway.

\item \emph{Repair step:} while the row replacement chooses the matrix and the row regardless of the conflicts they are involved in, the repair step chooses a conflicting subspace $E_i$ with probability proportional to its \emph{conflict cost} $C_i = \sum_{j \neq i} (2^{d_{ij}} - 1)$, and removes from $M_i$ the row that decreases the most the overlap with the other members of the family, giving again an $(m-1)$-dimensional subspace $W_i$. Then, a linear system is solved to find a new vector $v$ such that, when added to $W_i$, it does not create new conflicts with the other members. If the linear system has no solution, then the subspace is left unchanged. Hence, this step never increases the fitness.
\end{compactitem}
The following two operators are instead used with the spread-set encoding, the second one relying on the exact insertion procedure described in between:

\begin{compactitem}
\item \emph{Column replacement:} this operator is analogous to the row replacement, except that it acts on a square invertible matrix of the spread-set encoding. One column is selected at random, removed, and a new column is inserted by sampling it from the complement of the span of the reduced matrix, so that the matrix retains full rank (and thus invertibility). This operator was originally proposed in~\cite{mariot22}, and it is applied for each matrix/column pair in the family with probability $1/((t-2)m)$, again with at least one replacement.

\item \emph{Exact insertion:} given the matrices already present in the family (including the zero matrix, which represents $E_0$), this procedure searches for a new matrix $A$ such that its difference with each of them is invertible. The search exploits the fact that the $c$-th column of $A - B$ depends only on the $c$-th columns of $A$ and $B$. Thus, $A$ is built one column at a time, trying all $2^m$ possible vectors in random order for each column. A vector is discarded as soon as the columns chosen so far become linearly dependent in the difference with some matrix in the family, and when no vector is left, the search backtracks to the previous column. The procedure stops either with a full matrix $A$, or when all possibilities are exhausted (which proves that no such matrix exists), or after a budget of $10^6$ visited nodes.

\item \emph{Ruin-and-rebuild:} this operator, based on the ruin-and-recreate principle~\cite{schrimpf00}, removes from the family all members that are involved in a conflict, plus one further member chosen at random. Then, it rebuilds the removed positions in random order: for each of them, the exact insertion is run against all matrices present at that moment, including those already re-inserted. If a matrix is found, it takes that position; otherwise, the old matrix is restored.
\end{compactitem}

\subsection{Evolution Strategy and Constructive Baselines}
\label{sec:es}
\paragraph{Evolution Strategy.} In our experiments, we adopt a $(\mu+\lambda)$-ES: at each generation, the $\mu$ best individuals are selected as parents and generate $\lambda$ children in turn; the next population is formed by the parents and their children. With the subspace encoding, a child is created by applying first the row-replacement mutation operator and then the repair step; for the spread-set encoding, the column-replacement mutation is first applied, followed by the ruin-and-rebuild strategy. The initial population is composed of families of random full-rank generator matrices under the subspace encoding, or of random invertible matrices under the spread-set encoding. The ES stops as soon as it finds a partial spread, or when its budget is exhausted. We consider the variants $(3+7)$ and $(1+1)$. Compared to~\cite{carlet23}, where the $(\mu,\lambda)$ replacement without any repair gave the best results, we use the $(\mu+\lambda)$ replacement, since the former failed for $n=8$ in our preliminary runs, and we add the repair step, which speeds up convergence. For the second objective on the degree, besides the lexicographic variant of Section~\ref{sec:fitness}, we consider a \emph{quality-diversity} variant based on MAP-Elites~\cite{mouret15}: once the ES has found a first partial spread, an archive starts storing one partial spread (the \emph{elite}) for each observed value of $T$. At each step, an elite is chosen uniformly at random and produces one child as above. If the child is a partial spread, it enters the archive when the cell of its value of $T$ is empty; otherwise, it replaces the elite of that cell with probability $1/2$.
\paragraph{Constructive Baselines.} We compare the ES with two greedy heuristics that work on the spread-set encoding, namely the \emph{random greedy} and the \emph{look-ahead greedy}: both start from $E_\infty$ and $E_0$, and add one matrix at a time, until the target size is reached or no compatible matrix exists. In particular, the random greedy adds a random compatible matrix. The look-ahead greedy instead draws a set of compatible candidates and adds the one that preserves the compatibility of the largest part of a sample of compatible matrices. Both heuristics are restarted until they succeed or a time limit is reached, so that they are compared with the ES under the same running time.

\section{Experiments}
\label{sec:experiments}
We now discuss the experiments performed to evaluate the ES described in the previous section, starting with the experimental setup, and then presenting the results on the comparison of the subspace and spread-set encodings (E1), the ablation study and comparison with the constructive baselines (E2), the fitness landscape analysis together with the limit case $n=12$ (E3), the optimization of the algebraic degree (E4), and finally the analysis of the properties of the obtained $\mathcal{PS}$ functions (E5).

\subsection{Experimental Setup}
\label{subsec:setup}
In our experiments, we considered bent functions of $n=8$ and $n=10$ variables, both in the $\mathcal{PS}^-$ ($t=2^{m-1}$) and $\mathcal{PS}^+$ case ($t=2^{m-1}+1$), plus $n=12$ in the $\mathcal{PS}^-$ case as a limit instance. For each experimental configuration, we performed 100 independent runs, equally split into 25 runs on four computing nodes (one run per physical core), with disjoint seeds. The computing nodes used are IBM Power System AC922 with dual POWER9 CPUs, each with 32 physical cores, and all the software has been implemented in Java. Concerning the budgets of each run, we allocated: $10^6$ evaluations with a time cap of $1800$ seconds for E1 in both sizes and targets, plus two control configurations with $1800$ seconds and no limit on the evaluations; $300$ seconds for E2, again in both sizes and targets, and for the ES runs of E3; and $300$ seconds ($n=8$) and $1800$ seconds ($n=10$) for E4. For $n=12$, we performed 16 runs of 24 hours for each method. We adopted wall-clock time instead of evaluations for the comparison with the greedy baselines, since one of their steps is the insertion of a matrix and not the evaluation of a family.

For the ES parameters, we applied the repair step with probability 0.5, while the ruin-and-rebuild is done on every child, with a budget of $10^6$ nodes for each exact insertion. The look-ahead greedy uses 32 candidate matrices and a sample of 1000 compatible matrices.

To analyze the statistical significance of our results, the counts obtained by two methods (e.g., how many times they converge to a target partial spread in 100 runs) are compared with a two-sided Fisher's exact test~\cite{fisher34,agresti92}, and we report the resulting $p$-values. The thresholds have been determined by applying Bonferroni correction~\cite{bonferroni36,dunn61} on a significance level of $0.05$ within each experiment (except E1 and E5, where no test is reported).% yielding a corrected level of $0.025$ for E1 (2 tests), $0.0063$ for E2 (8 tests), $0.0167$ for E3 (3 tests, one of which is reported here) and $0.0071$ for E4 (7 tests). No test is involved in E5, since it is descriptive.

\subsection{E1: Encodings}
\label{subsec:rq1}
Table~\ref{tab:encodings} reports the number of successful runs of the $(3+7)$-ES under the two considered encodings; for the failed runs, the median number of conflicting pairs in the best family is given in parentheses. The two middle rows are control configurations for $n=10$: the subspace encoding with the same time budget as the spread-set one (1800 seconds, about $9 \cdot 10^7$ evaluations), and the spread-set encoding without the ruin-and-rebuild operator (about $2 \cdot 10^8$ evaluations).
\begin{table}[t]
\scriptsize
\centering
\caption{Runs solved out of 100 by the $(3+7)$-ES under the two encodings (E1).}
\label{tab:encodings}
\begin{tabular}{@{}llcccc@{}}
\toprule
 & & \multicolumn{2}{c}{$n=8$} & \multicolumn{2}{c}{$n=10$} \\
\cmidrule(lr){3-4} \cmidrule(l){5-6}
Encoding and second step & Budget & $\mathcal{PS}^-$ & $\mathcal{PS}^+$ & $\mathcal{PS}^-$ & $\mathcal{PS}^+$ \\
\midrule
Subspace, repair step            & $10^6$ evals          & 100 & 100 & 0 (9)  & 0 (13) \\
Subspace, repair step            & 1800 s                & --  & --  & 0 (5)  & 0 (8)  \\
Spread-set, no ruin-and-rebuild  & 1800 s                & --  & --  & 0 (8)  & 0 (12) \\
Spread-set, ruin-and-rebuild     & $10^6$ evals / 1800 s & 100 & 100 & 100    & 74 (2) \\
\bottomrule
\end{tabular}
\end{table}
For $n=8$, both encodings reach a full success rate on both targets, although the spread-set encoding needs a median of 17 evaluations, against 885 ($\mathcal{PS}^-$) and 9593 ($\mathcal{PS}^+$) for the subspace encoding. For $n=10$, only the spread-set encoding combined with ruin-and-rebuild converges to an optimal solution: in all runs for $\mathcal{PS}^-$ (median of 241 evaluations and 3.6 seconds), and in 74\% of the runs for $\mathcal{PS}^+$ within the 1800-second cap (median of about $5 \cdot 10^4$ evaluations and 687 seconds), where the failed runs end with a median of 2 conflicting pairs. All other configurations never find a partial spread. With 90 times more evaluations, the subspace encoding reduces its conflicts but remains far from a solution, and the spread-set encoding without ruin-and-rebuild ends where the subspace encoding does. Hence, neither component is sufficient on its own: ruin-and-rebuild is essential for the spread-set encoding to succeed, and the encoding is what makes the exact rebuilding possible.

\subsection{E2: Components and Baselines}
\label{subsec:rq2}
In the second experiment, we considered the ES with the most successful configuration from E1 (that is, spread-set encoding with ruin-and-rebuild), and performed an ablation study to investigate the importance of each component. Starting from the reference $(1+1)$-ES and $(3+7)$-ES, we consider variants without mutation (for the $(1+1)$-ES, this amounts to iterating ruin-and-rebuild alone), variants where the column replacement is substituted by bit flips on the entries of the matrices, either unconstrained or restricted to those that keep each matrix invertible, and a variant whose initial population is built by the look-ahead greedy heuristic. We also compare the ES with the restarted look-ahead and random greedy baselines. For $n=8$, every method solves all 100 runs on both targets; hence, Table~\ref{tab:components} reports the results for $n=10$ within 300 seconds, with the median time to success in seconds within parentheses.
\begin{table}[t]
\scriptsize
\centering
\caption{Runs solved out of 100 within 300 seconds, for $n=10$ (E2).}
\label{tab:components}
\begin{tabular}{@{}lcc@{}}
\toprule
Method & $\mathcal{PS}^-$ & $\mathcal{PS}^+$ \\
\midrule
Look-ahead greedy, restarted                  & 100 (1.8)  & 52 (122.7) \\
Random greedy, restarted                      &  99 (40.7) &  2 (139.1) \\
\midrule
$(1+1)$-ES                                    & 100 (3.4)  & 23 (109.0) \\
$(3+7)$-ES                                    & 100 (3.7)  & 20 (119.2) \\
\midrule
$(1+1)$-ES, no mutation                       &  97 (5.2)  &  4 (54.2)  \\
$(3+7)$-ES, no mutation                       &  98 (3.6)  &  0 (--)    \\
$(3+7)$-ES, bit flips keeping $A$ invertible  & 100 (3.5)  & 13 (220.2) \\
$(3+7)$-ES, unconstrained bit flips           & 100 (3.2)  & 17 (69.6)  \\
$(3+7)$-ES, look-ahead greedy initialization  & 100 (2.7)  & 13 (148.8) \\
\bottomrule
\end{tabular}
\end{table}
All variants solve between 97 and 100 runs for the $\mathcal{PS}^-$ target, so we focus on $\mathcal{PS}^+$, where we applied Fisher's exact test to 8 comparisons (corrected $\alpha=0.0063$). Among the baselines, the look-ahead greedy is far more successful than the random one ($p=3\cdot 10^{-17}$). It also outperforms the best ES variant, namely the $(1+1)$-ES ($p=4\cdot 10^{-5}$): on this instance, a restarted constructive heuristic is better than every ES variant, and initializing the ES with greedy families does not close the gap. Among the ES variants, removing the mutation significantly lowers the success rate, both for the $(3+7)$-ES ($p=7\cdot 10^{-7}$) and the $(1+1)$-ES ($p=1\cdot 10^{-4}$). The four remaining comparisons, which vary the population size, the type of mutation, and the initialization with respect to the reference ES, are not significant ($p\ge 0.25$). Hence, mutation is necessary for the harder target, while the population size, the mutation type, and a greedy initialization have no significant effect.

\begin{figure}[t]
\centering
\begin{minipage}[t]{0.485\textwidth}
  \centering
  \includegraphics[width=\linewidth]{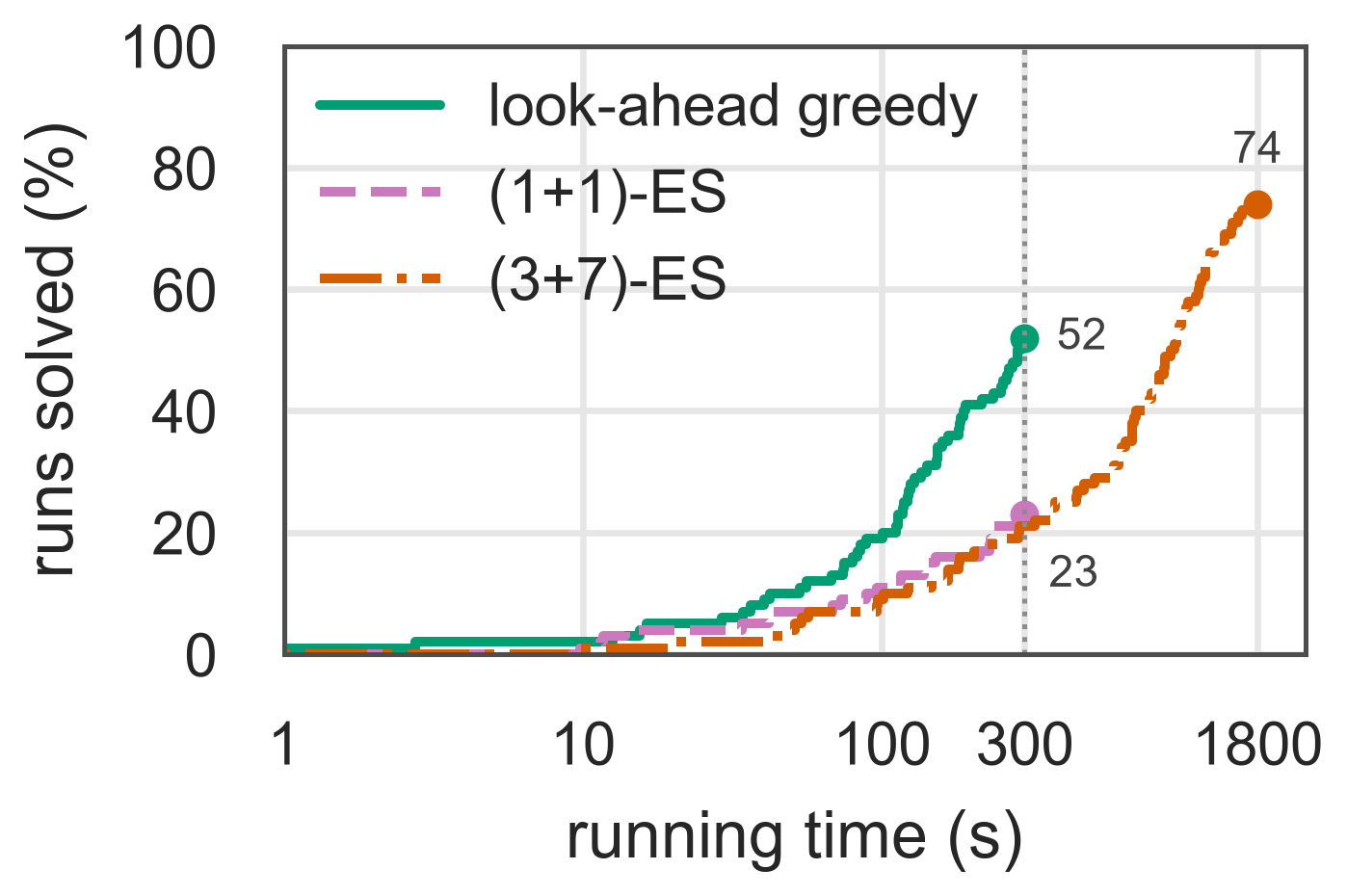}
  \caption{Runs solved over time for $n=10$ and $\mathcal{PS}^+$ ($t=17$), 100 runs per method.}
  \label{fig:runtime}
\end{minipage}\hfill
\begin{minipage}[t]{0.485\textwidth}
  \centering
  \includegraphics[width=\linewidth]{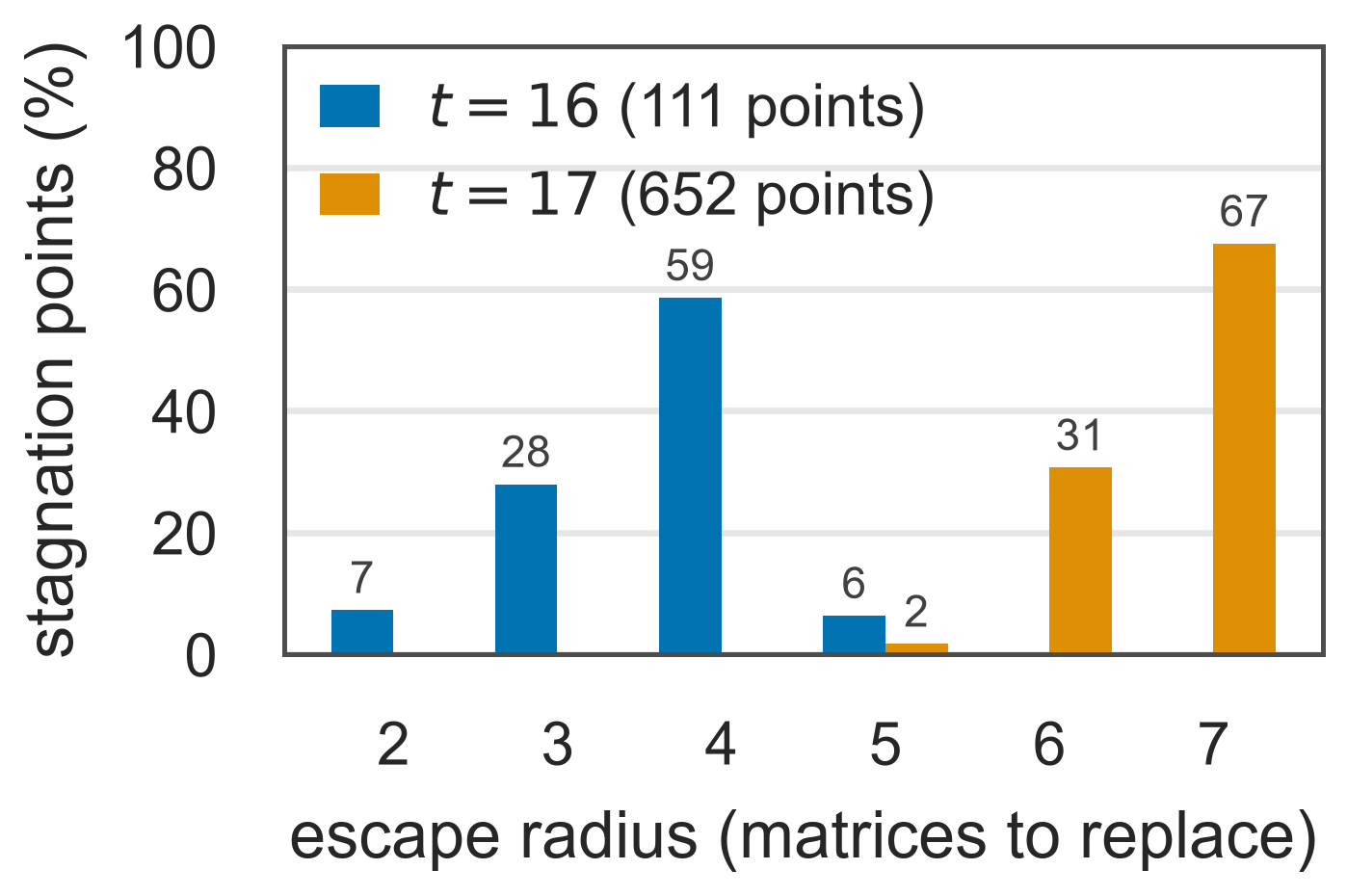}
  \caption{Escape radius of the stagnation points for $n=10$ ($t=16$: $\mathcal{PS}^-$, $t=17$: $\mathcal{PS}^+$).}
  \label{fig:radius}
\end{minipage}
\end{figure}

Figure~\ref{fig:runtime} plots the proportion of runs solved over time for $n=10$ in the $\mathcal{PS}^+$ case. The look-ahead greedy accumulates successes faster than both ES variants over the whole 300 seconds (dotted line). The curve of the $(3+7)$-ES is taken from the runs of E1, which use the same seeds and continue up to 1800 seconds (hence the small difference at 300 seconds with respect to Table~\ref{tab:components}): it keeps rising up to the 74\% of Table~\ref{tab:encodings}, showing that the two tables report the same configuration under different time limits.

\subsection{E3: Landscape Analysis and the Limit Case $n=12$}
\label{subsec:rq3}
To analyze the hardness of the $\mathcal{PS}^+$ target with respect to $\mathcal{PS}^-$ for $n=10$, in the third experiment we analyzed the stagnation points of the $(3+7)$-ES. We define a \emph{stagnation point} $P$ as the best family in the population when the best fitness has not improved for 10 generations. The \emph{escape radius} of $P$ is the minimum number $r$ of matrices whose removal leaves a conflict-free family that can be completed to a partial spread of the target size. We computed the escape radius exactly by trying all removal sets of increasing size up to $r=6$ for $\mathcal{PS}^-$ and $r=7$ for $\mathcal{PS}^+$. Figure~\ref{fig:radius} displays the distribution of the escape radius over the 111 stagnation points analyzed for $\mathcal{PS}^-$ and the 652 analyzed for $\mathcal{PS}^+$. The radius is mostly 4 (out of 14 matrices) in the former case, and mostly 7 (out of 15) in the latter. Considering only the first stagnation point of each run, the radius is at most 5 in all 68 runs that have one for $\mathcal{PS}^-$, and in 1 run out of 100 for $\mathcal{PS}^+$ ($p=6\cdot 10^{-47}$), including the runs that later succeed. Moreover, the minimal removal sets that can be completed are rarer for $\mathcal{PS}^+$ (about $0.4\%$ against $2\%$). Hence, the stagnation points of $\mathcal{PS}^+$ are farther from a partial spread. Remark that the ruin-and-rebuild strategy typically removes more matrices than the escape radius, since a single matrix is often in conflict with several others. However, the matrices are rebuilt one at a time in a random order; thus, the more matrices must be replaced at the same time, the less likely the rebuild step is successful.

For $n=12$, neither method found a partial spread in 16 runs of 24 hours. The restarted look-ahead greedy reached families of 23 subspaces out of 32 in every run, and never 24, over about 3770 attempts per run. The $(1+1)$-ES performed only about $4.8\cdot 10^4$ evaluations per run, ending with about 155 conflicting pairs: $84\%$ of its failed insertions were stopped by the node budget. Hence, this instance is out of reach for both approaches, because of the distance from the target and of the cost of each exact insertion.

\subsection{E4: Low-Degree Functions}
\label{subsec:rq4}
In the fourth experiment, we consider the search for $\mathcal{PS}^+$ functions with the additional objective of degree lower than $m$. To this end, we compare the lexicographic and quality-diversity variants of the ES with the restarted greedy heuristics, which record $T$ for every partial spread they build. For $n=8$ and 300 seconds, the runs reaching $T=0$ out of 100 are 90 and 84 for the quality-diversity $(1+1)$-ES and $(3+7)$-ES, 47 and 35 for the lexicographic ones, 27 for the look-ahead greedy and 2 for the random greedy. For the statistical analysis of the results, we performed 7 comparisons (corrected $\alpha=0.0071$). In general, the quality-diversity variants are significantly better than their lexicographic versions ($p\le 4\cdot 10^{-11}$) and the look-ahead greedy ($p\le 2\cdot 10^{-16}$). On the other hand, the two population sizes of the quality-diversity variant do not differ significantly ($p=0.29$). The lexicographic ES improves on the look-ahead greedy only in its $(1+1)$ form ($p=0.0052$, against $p=0.28$ for the $(3+7)$ form), and all its failed runs end at $T=1$: keeping only the best value of $T$ leads to a dead end. Overall, we obtained 285 distinct functions with $T=0$, of which 278 are of degree 3 and 7 are of degree 2, the latter all found by the quality-diversity variants. All these functions belong to the only two classes of degree lower than 4 that we found in the classification of $\mathcal{PS}$ functions of 8 variables by Langevin and Hou~\cite{langevin11}.

For $n=10$ and 1800 seconds, no method reached $T=0$, and the lowest value found is 99 out of 252 for the ES and the greedy heuristic alike. The median of the best $T$ per run is 116 for the look-ahead greedy, against 121 and 122 for the lexicographic and quality-diversity $(3+7)$-ES, which find a partial spread in only 72 and 73 runs out of 100.

\subsection{E5: Analysis of the Evolved Functions}
\label{subsec:rq5}
As a final experiment, we analyze the properties of the functions obtained in the first three experiments. Following the bounds of~\cite{weng07} recalled in Section~\ref{sec:background}, the 2-rank of an MM function is at most 30 for $n=8$ and 62 for $n=10$, while the 2-rank of a $\mathcal{PS}_{ap}$ function lies in $[30,42]$ and $[62,102]$, respectively. Moreover, the 2-rank of any bent function is at most 46 for $n=8$ and 112 for $n=10$, and it does not change by complementing the function~\cite{weng07}, so that these bounds apply to both targets. For $n=10$, all the functions that we found have a 2-rank of 112, the maximum possible value; hence, none of them is EA-equivalent to an MM or a $\mathcal{PS}_{ap}$ function. For $n=8$, the 2-ranks range between 24 and 46, thus reaching the upper bound also in this case. Almost all partial spreads for $n=10$ are non-extendable. Although a $\mathcal{PS}^+$ function can have degree lower than $m$ only if its partial spread is non-extendable, all these functions have degree 5.

To assess the diversity of the functions, we used as an EA-invariant (see, e.g.,~\cite{musukwa24}) the distribution of the Hamming weights of the second-order derivatives, defined as $x \mapsto f(x)\oplus f(x\oplus a)\oplus f(x\oplus b)\oplus f(x\oplus a\oplus b)$, over all 2-dimensional subspaces $\{\underline{0},a,b,a\oplus b\}$. For $n=10$, the 842 distinct $\mathcal{PS}^-$ functions and the 197 distinct $\mathcal{PS}^+$ functions all give different values, and are thus pairwise EA-inequivalent. For $n=8$, the 991 and 989 distinct functions belong to at least 866 and 835 classes, respectively.

We compared these functions with 300 bent functions for each of $n=8$ and $n=10$ evolved by GP~\cite{picek16} by maximizing the nonlinearity, with the following parameters: population size 500, maximum tree depth 5, function set $\{\textsc{and}, \textsc{or}, \textsc{not}, \textsc{xor}, \textsc{if}\}$, mutation probability $0.7$, and a steady-state 3-tournament selection. The functions obtained by GP are mostly of low degree (for $n=10$, 139 quadratic and 101 cubic functions out of 259 distinct ones), and their 2-ranks are at most 20 for $n=8$ and 32 for $n=10$. Hence, they are all below the lower bound for $\mathcal{PS}_{ap}$ functions, so that none of them is EA-equivalent to a function of this class, and there is no overlap with the 2-ranks of our functions. The two approaches thus reach very different regions of the set of bent functions: GP finds functions of low degree and low 2-rank, while the search on partial spreads gives functions of degree $m$ and high 2-rank, which is maximal for $n=10$.

\section{Conclusions}
\label{sec:conclusions}
We investigated the evolution of partial spreads for the construction of $\mathcal{PS}$ bent functions. The results show that the spread-set encoding combined with ruin-and-rebuild solves the instance $n=10$ for both targets, while neither of the two suffices alone. On the harder $\mathcal{PS}^+$ target, mutation is necessary and the population is not; a restarted look-ahead greedy heuristic outperforms every ES variant, and the stagnation points lie farther from a partial spread than for $\mathcal{PS}^-$. The functions found for $n=10$ are pairwise EA-inequivalent and have the maximum 2-rank, far from those evolved by GP. Concerning the degree, the quality-diversity ES finds $\mathcal{PS}^+$ functions of degree lower than $m$ for $n=8$ far more reliably than the greedy heuristics, while for $n=10$ no method found any, since reaching a partial spread is already the bottleneck. For future work, we plan to investigate stronger exact insertion methods for $n=12$, and symmetry-restricted encodings to search for $\mathcal{PS}^+$ functions of low degree for $n=10$.

\bibliographystyle{abbrv}
\bibliography{bibliography}

@book{carlet21,
    place={Cambridge},
    title={{B}oolean Functions for Cryptography and Coding Theory},
    publisher={Cambridge University Press},
    author={Carlet, Claude},
    year={2021}
}

@inproceedings{carlet23,
  author       = {Claude Carlet and
                  Luca Mariot and
                  Luca Manzoni and
                  Stjepan Picek},
  editor       = {Leslie P{\'{e}}rez C{\'{a}}ceres and
                  Thomas St{\"{u}}tzle},
  title        = {Evolutionary Strategies for the Design of Binary Linear Codes},
  booktitle    = {Evolutionary Computation in Combinatorial Optimization - 23rd European
                  Conference, EvoCOP 2023, Held as Part of EvoStar 2023, Brno, Czech
                  Republic, April 12-14, 2023, Proceedings},
  series       = {Lecture Notes in Computer Science},
  volume       = {13987},
  pages        = {114--129},
  publisher    = {Springer},
  year         = {2023}
}

@phdthesis{dillon74,
  title={Elementary {H}adamard difference sets},
  author={Dillon, John F},
  school={Univ. of Maryland},
  year={1974}
}

@article{gadouleau23,
  author       = {Maximilien Gadouleau and
                  Luca Mariot and
                  Stjepan Picek},
  title        = {Bent functions in the partial spread class generated by linear recurring
                  sequences},
  journal      = {Des. Codes Cryptogr.},
  volume       = {91},
  number       = {1},
  pages        = {63--82},
  year         = {2023}
}

@article{kirkman1850,
  title={Query {VI}},
  author={Kirkman, Thomas P.},
  journal={The Lady's and Gentleman's Diary},
  pages={48},
  year={1850}
}

@article{brown2009,
  title={{K}irkman's schoolgirls wearing hats and walking through fields of numbers},
  author={Brown, Ezra and Mellinger, Keith E},
  journal={Mathematics Magazine},
  volume={82},
  number={1},
  pages={3--15},
  year={2009},
  publisher={Taylor \& Francis}
}

@book{colbourn2019,
  title={Handbook of combinatorial designs},
  author={Colbourn, Charles J and Dinitz, Jeffrey H},
  year={2019},
  publisher={Chapman and Hall/CRC}
}

@article{djurasevic23,
  author       = {Marko Djurasevic and
                  Domagoj Jakobovic and
                  Luca Mariot and
                  Stjepan Picek},
  title        = {A survey of metaheuristic algorithms for the design of cryptographic
                  {B}oolean functions},
  journal      = {Cryptogr. Commun.},
  volume       = {15},
  number       = {6},
  pages        = {1171--1197},
  year         = {2023}
}

@article{picek16,
	author       = {Stjepan Picek and
	Domagoj Jakobovic and
	Julian F. Miller and
	Lejla Batina and
	Marko Cupic},
	title        = {Cryptographic {B}oolean functions: One output, many design criteria},
	journal      = {Appl. Soft Comput.},
	volume       = {40},
	pages        = {635--653},
	year         = {2016}
}

@article{rothaus76,
  author       = {Oscar S. Rothaus},
  title        = {On "Bent" Functions},
  journal      = {J. Comb. Theory {A}},
  volume       = {20},
  number       = {3},
  pages        = {300--305},
  year         = {1976}
}

@article{weng07,
  author       = {Guobiao Weng and
                  Rongquan Feng and
                  Weisheng Qiu},
  title        = {On the ranks of bent functions},
  journal      = {Finite Fields Their Appl.},
  volume       = {13},
  number       = {4},
  pages        = {1096--1116},
  year         = {2007}
}

@article{mcfarland73,
  author       = {Robert L. McFarland},
  title        = {A Family of Difference Sets in Non-cyclic Groups},
  journal      = {J. Comb. Theory {A}},
  volume       = {15},
  number       = {1},
  pages        = {1--10},
  year         = {1973}
}

@inproceedings{fuller03,
  author    = {Joanne Fuller and
               Ed Dawson and
               William Millan},
  title     = {Evolutionary generation of bent functions for cryptography},
  booktitle = {Proceedings of the {IEEE} Congress on Evolutionary Computation, {CEC}
               2003, Canberra, Australia, December 8-12, 2003},
  pages     = {1655--1661},
  publisher = {{IEEE}},
  year      = {2003}
}

@inproceedings{clark03,
  author       = {John A. Clark and
                  Jeremy L. Jacob and
                  Subhamoy Maitra and
                  Pantelimon Stanica},
  title        = {Almost {B}oolean functions: the design of {B}oolean functions by spectral
                  inversion},
  booktitle    = {Proceedings of the {IEEE} Congress on Evolutionary Computation, {CEC}
                  2003, Canberra, Australia, December 8-12, 2003},
  pages        = {2173--2180},
  publisher    = {{IEEE}},
  year         = {2003}
}

@inproceedings{StanicaMC04,
  author    = {Pantelimon Stanica and
               Subhamoy Maitra and
               John A. Clark},
  editor    = {Bimal K. Roy and
               Willi Meier},
  title     = {Results on Rotation Symmetric Bent and Correlation Immune {B}oolean
               Functions},
  booktitle = {Fast Software Encryption, 11th International Workshop, {FSE} 2004,
               Delhi, India, February 5-7, 2004, Revised Papers},
  series    = {Lecture Notes in Computer Science},
  volume    = {3017},
  pages     = {161--177},
  publisher = {Springer},
  year      = {2004},
  address={Berlin, Heidelberg}
}

@inproceedings{hrbacek14,
    author="Hrbacek, Radek and Dvorak, Vaclav",
    editor="Bartz-Beielstein, Thomas and Branke, J{\"u}rgen and Filipi{\v{c}}, Bogdan and Smith, Jim",
    title="Bent Function Synthesis by Means of Cartesian Genetic Programming",
    booktitle="Parallel Problem Solving from Nature -- PPSN XIII",
    year="2014",
    publisher="Springer International Publishing",
    address="Cham",
    pages="414--423"
}

@inproceedings{picekJ16,
  author       = {Stjepan Picek and
                  Domagoj Jakobovic},
  editor       = {Tobias Friedrich and
                  Frank Neumann and
                  Andrew M. Sutton},
  title        = {Evolving Algebraic Constructions for Designing Bent {B}oolean Functions},
  booktitle    = {Proceedings of the 2016 on Genetic and Evolutionary Computation Conference,
                  Denver, CO, USA, July 20 - 24, 2016},
  pages        = {781--788},
  publisher    = {{ACM}},
  year         = {2016}
}

@inproceedings{picek18,
  author       = {Stjepan Picek and
                  Karlo Knezevic and
                  Luca Mariot and
                  Domagoj Jakobovic and
                  Alberto Leporati},
  title        = {Evolving Bent Quaternary Functions},
  booktitle    = {2018 {IEEE} Congress on Evolutionary Computation, {CEC} 2018, Rio
                  de Janeiro, Brazil, July 8-13, 2018},
  pages        = {1--8},
  publisher    = {{IEEE}},
  year         = {2018}
}

@inproceedings{mariot19,
  author       = {Luca Mariot and
                  Domagoj Jakobovic and
                  Alberto Leporati and
                  Stjepan Picek},
  editor       = {Luk{\'{a}}s Sekanina and
                  Ting Hu and
                  Nuno Louren{\c{c}}o and
                  Hendrik Richter and
                  Pablo Garc{\'{\i}}a{-}S{\'{a}}nchez},
  title        = {Hyper-bent {B}oolean Functions and Evolutionary Algorithms},
  booktitle    = {Genetic Programming - 22nd European Conference, EuroGP 2019, Held
                  as Part of EvoStar 2019, Leipzig, Germany, April 24-26, 2019, Proceedings},
  series       = {Lecture Notes in Computer Science},
  volume       = {11451},
  pages        = {262--277},
  publisher    = {Springer},
  year         = {2019}
}

@article{manzoni20,
  author       = {Luca Manzoni and
                  Luca Mariot and
                  Eva Tuba},
  title        = {Balanced crossover operators in Genetic Algorithms},
  journal      = {Swarm Evol. Comput.},
  volume       = {54},
  pages        = {100646},
  year         = {2020}
}

@article{mariot22,
  author       = {Luca Mariot and
                  Martina Saletta and
                  Alberto Leporati and
                  Luca Manzoni},
  title        = {Heuristic search of (semi-)bent functions based on cellular automata},
  journal      = {Nat. Comput.},
  volume       = {21},
  number       = {3},
  pages        = {377--391},
  year         = {2022}
}

@inproceedings{carlet24a,
  author       = {Claude Carlet and
                  Marko Durasevic and
                  Domagoj Jakobovic and
                  Luca Mariot and
                  Stjepan Picek},
  editor       = {Mario Giacobini and
                  Bing Xue and
                  Luca Manzoni},
  title        = {Look into the Mirror: Evolving Self-dual Bent {B}oolean Functions},
  booktitle    = {Genetic Programming - 27th European Conference, EuroGP 2024, Held
                  as Part of EvoStar 2024, Aberystwyth, UK, April 3-5, 2024, Proceedings},
  series       = {Lecture Notes in Computer Science},
  volume       = {14631},
  pages        = {161--175},
  publisher    = {Springer},
  year         = {2024}
}

@inproceedings{carlet26a,
  author       = {Claude Carlet and
                  Marko {\DH}urasevic and
                  Domagoj Jakobovic and
                  Luca Mariot and
                  Stjepan Picek and
                  Alexandr Polujan},
  editor       = {Pablo Garc{\'{\i}}a{-}S{\'{a}}nchez and
                  Josefa D{\'{\i}}az {\'{A}}lvarez and
                  Aidan Murphy},
  title        = {On Counts and Densities of Homogeneous Bent Functions: An Evolutionary
                  Approach},
  booktitle    = {Applications of Evolutionary Computation - 29th European Conference,
                  EvoApplications 2026, Held as Part of EvoStar 2026, Toulouse, France,
                  April 8-10, 2026, Proceedings, Part {I}},
  series       = {Lecture Notes in Computer Science},
  volume       = {16524},
  pages        = {216--232},
  publisher    = {Springer},
  year         = {2026}
}

@inproceedings{carlet26b,
  author       = {Claude Carlet and
                  Marko Durasevic and
                  Ermes Franch and
                  Domagoj Jakobovic and
                  Luca Mariot and
                  Stjepan Picek},
  editor       = {Leonardo Trujillo and
                  Ting Hu},
  title        = {{N}ega{B}ent, No Regrets: Evolving Spectrally Flat {B}oolean Functions},
  booktitle    = {Proceedings of the Genetic and Evolutionary Computation Conference
                  Companion, {GECCO} 2026, Centro Internacional de Convenciones CIC-ANDE,
                  San Jose, Costa Rica, July 13-17, 2026},
  pages        = {589--592},
  publisher    = {{ACM}},
  year         = {2026}
}

@book{mullen13,
	editor    = {Gary L. Mullen and
	Daniel Panario},
	title     = {Handbook of Finite Fields},
	series    = {Discrete mathematics and its applications},
	publisher = {{CRC} Press},
	year      = {2013}
}

@article{bruck64,
  title={The construction of translation planes from projective spaces},
  author={Bruck, Richard Hubert and Bose, Raj Chandra},
  journal={Journal of Algebra},
  volume={1},
  number={1},
  pages={85--102},
  year={1964},
  publisher={Elsevier}
}

@article{rao46,
  title={Difference sets and combinatorial arrangements derivable from finite geometries},
  author={Rao, C. Radhakrishna},
  journal={Proceedings of the Indian National Science Academy-Part A: Physical Sciences},
  volume={12},
  number={3},
  pages={123--135},
  year={1946},
  publisher={Indian National Science Academy}
}

@article{schrimpf00,
  title={Record breaking optimization results using the ruin and recreate principle},
  author={Schrimpf, Gerhard and Schneider, Johannes and Stamm-Wilbrandt, Hermann and Dueck, Gunter},
  journal={Journal of Computational Physics},
  volume={159},
  number={2},
  pages={139--171},
  year={2000},
  publisher={Elsevier}
}

@article{mouret15,
  title={Illuminating search spaces by mapping elites},
  author={Mouret, Jean-Baptiste and Clune, Jeff},
  journal={arXiv preprint arXiv:1504.04909},
  year={2015}
}

@book{fisher34,
  title={Statistical Methods for Research Workers},
  author={Fisher, Ronald Aylmer},
  edition={5},
  publisher={Oliver and Boyd},
  address={Edinburgh},
  year={1934}
}

@article{agresti92,
  title={A survey of exact inference for contingency tables},
  author={Agresti, Alan},
  journal={Statistical science},
  volume={7},
  number={1},
  pages={131--153},
  year={1992},
  publisher={Institute of Mathematical Statistics}
}

@article{bonferroni36,
  title={Teoria statistica delle classi e calcolo delle probabilit{\`a}},
  author={Bonferroni, Carlo},
  journal={Pubblicazioni del R. Istituto Superiore di Scienze Economiche e Commerciali di Firenze},
  volume={8},
  pages={3--62},
  year={1936}
}

@article{dunn61,
  title={Multiple comparisons among means},
  author={Dunn, Olive Jean},
  journal={Journal of the American statistical association},
  volume={56},
  number={293},
  pages={52--64},
  year={1961},
  publisher={Taylor \& Francis}
}

@article{langevin11,
  author       = {Philippe Langevin and
                  Xiang{-}dong Hou},
  title        = {Counting Partial Spread Functions in Eight Variables},
  journal      = {{IEEE} Trans. Inf. Theory},
  volume       = {57},
  number       = {4},
  pages        = {2263--2269},
  year         = {2011}
}

@article{musukwa24,
  title={On Second-Order Derivatives of {B}oolean Functions and Cubic {APN} Permutations in Even Dimension},
  author={Musukwa, Augustine and Sala, Massimiliano and Villa, Irene and Zaninelli, Marco},
  journal={Mediterranean Journal of Mathematics},
  volume={21},
  number={3},
  pages={116},
  year={2024},
  publisher={Springer}
}

\end{document}